\documentclass{article}
\usepackage{iclr2026_conference,times}

\usepackage{amsmath,amsfonts,bm}

\def\eqref#1{equation~\ref{#1}}

\def\1{\bm{1}}

\DeclareMathAlphabet{\mathsfit}{\encodingdefault}{\sfdefault}{m}{sl}
\SetMathAlphabet{\mathsfit}{bold}{\encodingdefault}{\sfdefault}{bx}{n}

\usepackage{hyperref}
\usepackage{url}
\usepackage{graphicx}
\usepackage{booktabs}
\usepackage{amsmath,amssymb}
\usepackage{multirow}
\usepackage{xcolor}
\usepackage{enumitem}

\title{ChaosBench-Logic v2: Evaluating LLM Logical Reasoning\\over Dynamical Systems at Scale}

\author{
Noel Thomas\\
Mohamed bin Zayed University of Artificial Intelligence\\
Abu Dhabi, UAE\\
\texttt{noel.thomas@mbzuai.ac.ae}
}

\iclrfinalcopy

\begin{document}

\maketitle

\begin{abstract}
Standard accuracy on binary reasoning benchmarks hides critical failure modes: prior collapse, inconsistency under paraphrase, and inability to reason about parameter-dependent dynamics.
We present ChaosBench-Logic v2, a 40,886-question benchmark over 165 dynamical systems with 27~FOL predicates and 78~axiom edges, together with CARE (Calibration- and Adversarial-Robust Evaluation), a protocol that surfaces these pathologies.
Evaluating 14~models, we find that regime transition reasoning remains near-random (MCC\,=\,0.05) even for frontier models, while FOL deduction with given premises reaches MCC\,=\,0.52; per-family decomposition shows the proprietary advantage concentrates on cross-indicator (+0.40) and consistency tasks, while open-source Qwen\,2.5-32B dominates indicator diagnostics (0.91 vs.\ 0.45).
Two models exhibit negative MCC on bifurcation questions, confirmed as systematic anti-correlation via confusion matrix analysis.
\end{abstract}

\section{Introduction}

Large language models achieve strong performance on mathematical and logical reasoning benchmarks~\citep{wei2022cot, cobbe2021gsm8k}, yet their capacity for \emph{logically consistent} reasoning over scientific domains remains poorly understood.
Dynamical systems present a particularly demanding testbed: chaos is deterministic but not random, exhibits sensitive dependence on initial conditions, and requires positive Lyapunov exponents~\citep{strogatz2018nonlinear}.
These formal distinctions must be maintained across multi-step inferences, a requirement that probes deeper than pattern matching.

Existing benchmarks target mathematical problem-solving~\citep{hendrycks2021math, cobbe2021gsm8k}, scientific QA~\citep{clark2018arc, wang2023scibench}, logical reasoning~\citep{liu2020logiqa, han2022folio}, or synthetic FOL reasoning~\citep{saparov2023prontoqa}, but none combine (i)~a domain-specific FOL ontology with (ii)~ground-truth labels derived from axiom entailment over (iii)~real scientific systems at scale.

We introduce ChaosBench-Logic v2, a 66$\times$ scale-up of v1~\citep{thomas2026chaosbench}: from 621 to 40,886 questions, 30 to 165 systems, and 7 to 11 task families.
Our contributions:
\begin{enumerate}[nosep]
  \item \textbf{Benchmark.} 40,886 questions, 11 task families, 27 predicates, 78 FOL axiom edges, 165 dynamical systems (135 from \texttt{dysts}~\citep{gilpin2021chaos}).
  \item \textbf{CARE protocol.} A calibration- and robustness-aware evaluation framework (MCC, macro-family MCC, calibration diagnostics, consistency, coverage) that exposes failure modes hidden by accuracy.
  \item \textbf{Diagnostic findings.} A knowledge-type boundary between rule-following and parameter-dependent reasoning, per-family decomposition of the proprietary--OSS gap, and systematic prediction biases including negative MCC.
\end{enumerate}

\section{Related Work}

\paragraph{LLM reasoning benchmarks.}
GSM8K~\citep{cobbe2021gsm8k} and MATH~\citep{hendrycks2021math} test mathematical reasoning; ARC~\citep{clark2018arc} tests science QA; LogiQA~\citep{liu2020logiqa} and FOLIO~\citep{han2022folio} test logical reasoning; BIG-Bench~\citep{srivastava2023bigbench} includes some logical tasks.
PrOntoQA~\citep{saparov2023prontoqa} is closest to our work: it tests compositional FOL reasoning over synthetic ontologies, finding that LLMs struggle with longer inference chains.
LogicBench~\citep{parmar2024logicbench} evaluates 25 logical reasoning patterns and confirms failures on complex reasoning with negation.
Our benchmark differs by grounding FOL axioms in a real scientific domain where ground-truth labels derive from physical properties rather than synthetic constructions.

\paragraph{Consistency and robustness.}
TruthfulQA~\citep{lin2022truthfulqa} measures truthfulness; self-consistency decoding~\citep{wang2023selfconsistency} improves CoT reliability; ReClor~\citep{yu2020reclor} tests logical reading comprehension.
Our consistency\_paraphrase and perturbation families extend these ideas to a scientific domain with formal ground truth.

\paragraph{Scientific reasoning and dynamical systems.}
SciBench~\citep{wang2023scibench} evaluates college-level scientific problem-solving.
The \texttt{dysts} library~\citep{gilpin2021chaos} provides 135 standardized dynamical systems originally designed for forecasting benchmarks; we build on it for system diversity, preserving provenance and verifying predicate annotations against our axiom system.
ChaosBench-Logic v1~\citep{thomas2026chaosbench} introduced a 621-question benchmark; we scale by two orders of magnitude.

\section{Benchmark Design}

\subsection{Ontology}

The benchmark is grounded in a first-order logic ontology of 27 unary predicates in three tiers: 11 \textbf{core predicates} characterizing dynamical regimes (\textit{Chaotic}, \textit{Deterministic}, \textit{PosLyap}, \textit{Sensitive}, \textit{StrangeAttr}, \textit{PointUnpredictable}, \textit{StatPredictable}, \textit{QuasiPeriodic}, \textit{Random}, \textit{FixedPointAttr}, \textit{Periodic}), 4 \textbf{topological predicates} (\textit{Dissipative}, \textit{Bounded}, \textit{Mixing}, \textit{Ergodic}), and 12 \textbf{structural predicates} (\textit{HyperChaotic}, \textit{Conservative}, \textit{ContinuousTime}, \textit{DiscreteTime}, etc.).

These predicates are connected by \textbf{78 directed axiom edges} (31 implication, 47 exclusion), enabling reasoning chains of up to 5--6 hops.
For example, the Chaotic predicate entails:
\begin{align}
\forall s:\; \text{Chaotic}(s) &\Rightarrow \text{Deterministic}(s) \land \text{PosLyap}(s) \land \text{Sensitive}(s) \land \text{Mixing}(s) \nonumber \\
&\quad \land\; \neg\text{Random}(s) \land \neg\text{Periodic}(s) \land \neg\text{QuasiPeriodic}(s) \label{eq:chaos-axiom}
\end{align}
The full specification is in Appendix~\ref{app:axioms}.

\subsection{Systems}

The benchmark covers 165 dynamical systems:\footnote{Regime transition and FOL inference families additionally use synthetic parameterizations not counted in this total.} 30 manually curated (Lorenz-63~\citep{lorenz1963deterministic}, R\"ossler, H\'enon, logistic map, Brusselator, Ornstein-Uhlenbeck, etc.) and 135 from the \texttt{dysts} library~\citep{gilpin2021chaos}.
Each system carries ground-truth values for all 27 predicates, verified against the axiom system.

\subsection{Task Families}

\begin{table}[t]
\caption{Dataset composition by task family, ordered by difficulty. $N < 100$ families are interpreted qualitatively.}
\label{tab:families}
\centering
\small
\begin{tabular}{lrl}
\toprule
\textbf{Family} & \textbf{N} & \textbf{Description} \\
\midrule
regime\_transition & 68 & Bifurcation-dependent behavior \\
cross\_indicator & 67 & Multi-indicator reasoning \\
consistency\_paraphrase & 4,139 & Linguistic variation stability \\
perturbation & 1,994 & Parameter perturbation robustness \\
atomic & 25,307 & Single-predicate queries \\
adversarial\_nearmiss & 478 & Near-miss misconceptions \\
adversarial\_misleading & 500 & Misleading cue probes \\
multi\_hop & 6,000 & 2--6 step inference chains \\
fol\_inference & 1,758 & FOL deduction from premises \\
indicator\_diagnostic & 530 & Chaos indicator interpretation \\
extended\_systems & 45 & Factual recall (ceiling check) \\
\midrule
\textbf{Total} & \textbf{40,886} & \\
\bottomrule
\end{tabular}
\end{table}

\textbf{Multi-hop} questions chain 2--6 steps through the axiom graph (e.g., ``FluidTrampoline is strongly mixing $\Rightarrow$ weakly mixing $\Rightarrow$ ergodic $\Rightarrow$ bounded. Is it bounded?'').
\textbf{Regime transition} questions require specific bifurcation thresholds (e.g., ``At $\alpha$=15.6, is Chua's circuit chaotic?'').
\textbf{FOL inference} questions present premises for deductive conclusions.
\textbf{Adversarial} questions include misleading premises irrelevant to the answer.
\textbf{Indicator diagnostic} questions require interpreting numerical chaos indicators (0-1 test~\citep{gottwald2009zerone}, permutation entropy~\citep{bandt2002permutation}, MEGNO~\citep{cincotta2000megno}).
Representative examples with model predictions are in Appendix~\ref{app:examples}.

\subsection{Evaluation Protocol: CARE}

Standard accuracy on binary classification benchmarks can be misleading.
Since $\text{Acc} = p \cdot \text{TPR} + (1-p) \cdot \text{TNR}$ where $p$ is the TRUE prevalence, a model with high TNR but low TPR inflates accuracy by exploiting class priors.
LLaMA\,3.1-8B illustrates this: its TPR\,=\,0.32 and TNR\,=\,0.88 yield 60.2\% accuracy (vs.\ 50.5\% for always-FALSE), but MCC\,=\,0.24 and balanced accuracy\,=\,0.60 correctly signal near-chance performance.
To surface such pathologies, we propose \textbf{CARE} (\textbf{C}alibration- and \textbf{A}dversarial-\textbf{R}obust \textbf{E}valuation), a protocol for reasoning benchmarks.

CARE reports five diagnostics:
\begin{enumerate}[nosep]
\item \textbf{MCC} (primary)~\citep{matthews1975mcc, chicco2020mcc}: penalizes prior collapse and is invariant to class balance. Ranges from $-1$ (anti-correlation) through $0$ (random) to $+1$ (perfect):
\begin{equation}
\text{MCC} = \frac{TP \cdot TN - FP \cdot FN}{\sqrt{(TP+FP)(TP+FN)(TN+FP)(TN+FN)}}
\end{equation}
\item \textbf{Macro-family MCC}: mean MCC across task families, preventing dominant families (atomic: 62\% of questions) from masking failures on hard families.
\item \textbf{Calibration}: predicted TRUE rate vs.\ ground-truth TRUE rate. Flags prior collapse (e.g., LLaMA\,3.1-8B predicts TRUE only 21.6\% vs.\ 49.5\% ground truth).
\item \textbf{Consistency}: MCC on \textit{consistency\_paraphrase} and \textit{perturbation} families, measuring whether predictions survive surface-level variation.
\item \textbf{Coverage}: invalid rate (unparseable responses counted as incorrect) and per-family coverage, flagging instruction-following failures.
\end{enumerate}

Table~\ref{tab:care} demonstrates CARE on four models.
Accuracy ranks LLaMA\,3.1-8B at 60.2\% (above chance), but CARE flags prior collapse (21.6\% predicted TRUE vs.\ 49.5\% ground truth) and asymmetric recall (TPR\,=\,0.32, TNR\,=\,0.88): the model achieves accuracy by defaulting to FALSE, not by reasoning.
Mistral-7B's 61.3\% accuracy masks a 1.1\% invalid rate and consistency MCC below 0.30.
Only Claude Sonnet 4.6 triggers no CARE flags.

\begin{table}[t]
\caption{CARE diagnostics for four models. Flags: \textsc{pc}\,=\,prior collapse ($|\text{pred TRUE\%} - 49.5| > 10$); \textsc{ar}\,=\,asymmetric recall ($|\text{TPR}-\text{TNR}| > 0.3$); \textsc{cv}\,=\,coverage ($>$0.5\% invalid); \textsc{ic}\,=\,inconsistent (consistency MCC\,$<$\,0.30).}
\label{tab:care}
\centering
\small
\begin{tabular}{lcccccl}
\toprule
\textbf{Model} & \textbf{Acc} & \textbf{MCC} & \textbf{TPR} & \textbf{TNR} & \textbf{Pred T\%} & \textbf{Flags} \\
\midrule
Claude Sonnet 4.6 & 79.8 & 0.601 & 0.72 & 0.87 & 42.5 & (none) \\
Qwen\,2.5-32B & 73.8 & 0.478 & 0.68 & 0.79 & 44.4 & \textsc{ic} \\
LLaMA\,3.1-8B & 60.2 & 0.240 & 0.32 & 0.88 & 21.6 & \textsc{pc, ar, ic} \\
Mistral-7B & 61.3 & 0.228 & 0.67 & 0.55 & 56.2 & \textsc{cv, ic} \\
\bottomrule
\end{tabular}
\end{table}

All standard models use temperature\,=\,0; reasoning models (o3-mini, GPT-5.2) do not accept a temperature parameter and are deterministic by design (Appendix~\ref{app:prompt}).
Most models receive max\_tokens\,=\,16; reasoning models and Gemini use 1024 because their architectures consume output tokens for internal reasoning.

\section{Experiments}

We evaluate 14 models: 7 proprietary (Claude Sonnet 4.6, GPT-5.2, GPT-4o, GPT-4o-mini, o3-mini, Gemini 2.5 Flash, DeepSeek-Chat) and 7 open-source (Qwen\,2.5-\{7B,14B,32B\}, LLaMA\,3.3-70B, LLaMA\,3.1-8B, Gemma2-9B, Mistral-7B) served via Ollama.
Ten models complete the full dataset (N\,=\,40,886); four use subsets (5k or 1k) due to compute constraints, reported in a separate table to avoid cross-$N$ ranking.

\section{Results}

\subsection{Overall Performance}

\begin{table}[t]
\caption{Full canonical (N\,=\,40,886), ranked by MCC.}
\label{tab:leaderboard}
\centering
\small
\begin{tabular}{llcc}
\toprule
\textbf{Model} & \textbf{Type} & \textbf{Bal.\ Acc} & \textbf{MCC} \\
\midrule
Claude Sonnet 4.6 & Prop. & 0.797 & \textbf{0.601} \\
GPT-5.2 & Prop. & 0.747 & 0.509 \\
Qwen\,2.5-32B & OSS & 0.738 & 0.478 \\
DeepSeek-Chat & Prop. & 0.724 & 0.469 \\
Gemini 2.5 Flash & Prop. & 0.718 & 0.458 \\
GPT-4o & Prop. & 0.721 & 0.450 \\
Qwen\,2.5-14B & OSS & 0.711 & 0.426 \\
LLaMA\,3.3-70B & OSS & 0.681 & 0.373 \\
LLaMA\,3.1-8B & OSS & 0.599 & 0.240 \\
Mistral-7B & OSS & 0.613 & 0.228 \\
\bottomrule
\end{tabular}
\end{table}

Claude Sonnet 4.6 leads with MCC\,=\,0.601 (Table~\ref{tab:leaderboard}).
The proprietary--OSS gap is 0.12 MCC, but Qwen\,2.5-32B (0.478) outperforms GPT-4o (0.450) and Gemini 2.5 Flash (0.458).
Subset evaluations (o3-mini MCC\,=\,0.608 on 5k; Gemma2-9B 0.280 on 5k; GPT-4o-mini 0.272 on 1k; Qwen\,2.5-7B 0.268 on 1k) are reported in Appendix~\ref{app:subset}.

\subsection{Task Family Hardness}

\begin{figure}[t]
\centering
\includegraphics[width=\linewidth]{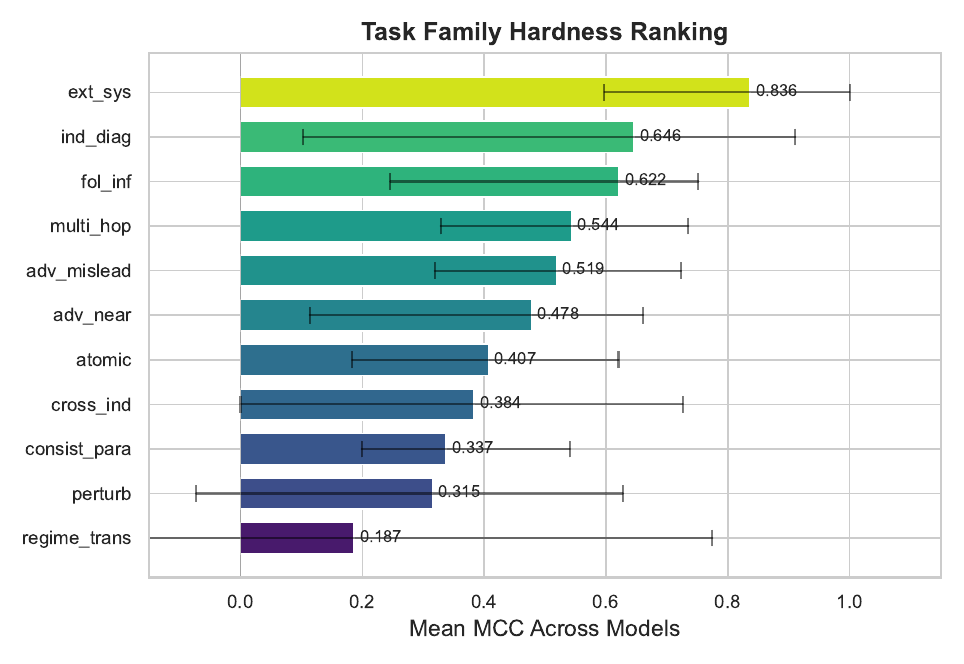}
\caption{Mean MCC by task family across 10 full-canonical models. Error bars: min--max.}
\label{fig:hardness}
\end{figure}

Figure~\ref{fig:hardness} ranks families by mean MCC.
\textbf{Easy} (MCC\,$>$\,0.5): \textit{extended\_systems} (0.81, ceiling effect), \textit{indicator\_diagnostic} (0.59), \textit{fol\_inference} (0.52).
\textbf{Medium} (0.25--0.5): \textit{multi\_hop} (0.48), \textit{adversarial} families (0.44--0.45), \textit{atomic} (0.32).
\textbf{Hard} ($<$\,0.25): \textit{perturbation} (0.26), \textit{consistency\_paraphrase} (0.25), \textit{cross\_indicator} (0.18), \textit{regime\_transition} (0.05).
Regime transition is near-random for all models: these questions require specific bifurcation thresholds (e.g., logistic map at $r \approx 3.57$) not recoverable from logical rules.

\subsection{Per-Model Family Analysis}

\begin{figure}[t]
\centering
\includegraphics[width=\linewidth]{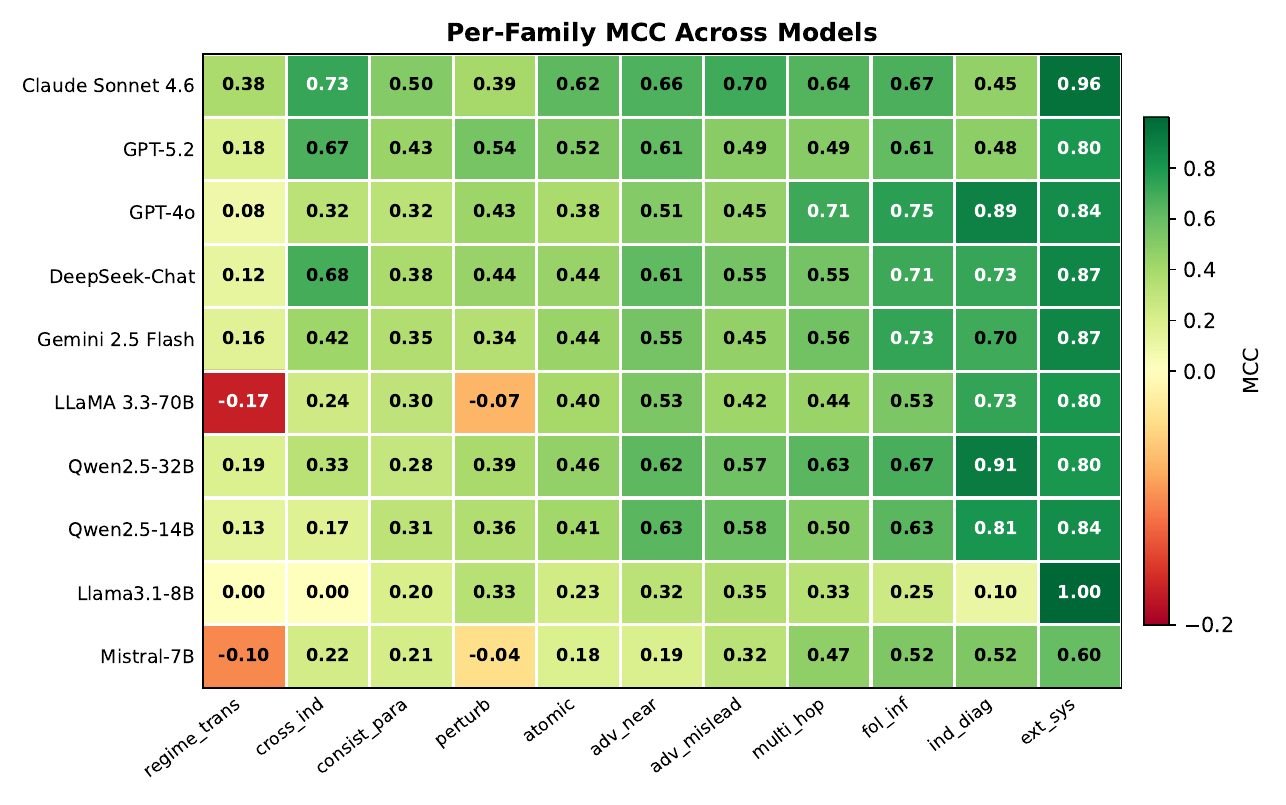}
\caption{Per-family MCC for 10 models. Families ordered by hardness (left\,=\,hardest). Red cells indicate negative MCC (anti-correlation).}
\label{fig:heatmap}
\end{figure}

Figure~\ref{fig:heatmap} reveals that family-level performance is not monotonic with overall MCC.
Qwen\,2.5-32B achieves MCC\,=\,0.91 on \textit{indicator\_diagnostic} (exceeding GPT-4o at 0.89 and Claude Sonnet at 0.45), while Claude Sonnet leads on \textit{multi\_hop} (0.64) and \textit{atomic} (0.62).
LLaMA\,3.1-8B scores perfectly on \textit{extended\_systems} (45 factual-recall questions, ceiling effect) but near-zero on \textit{cross\_indicator}.

Two models produce \textbf{negative MCC} on regime\_transition: LLaMA\,3.3-70B ($-0.17$; TP\,=\,9, FP\,=\,17, TN\,=\,20, FN\,=\,22; balanced accuracy 0.42) and Mistral-7B ($-0.10$).
Negative MCC means systematic anti-correlation: these models have learned heuristics that are reliably wrong on bifurcation questions.
Confusion matrices are in Appendix~\ref{app:confusion}.

\subsection{Where the Proprietary Advantage Concentrates}

\begin{figure}[t]
\centering
\includegraphics[width=\linewidth]{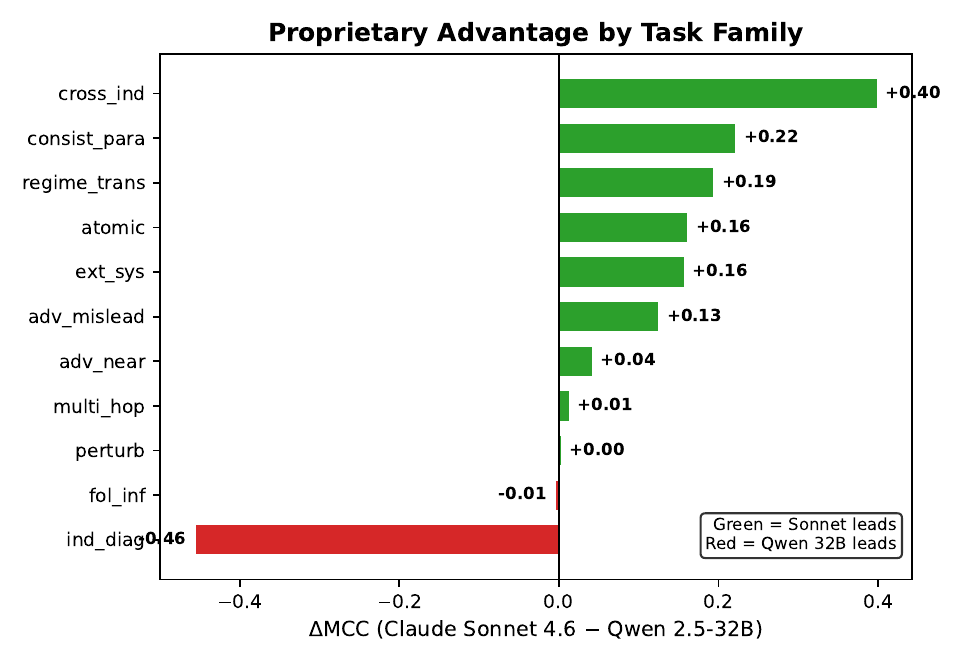}
\caption{Per-family $\Delta$MCC (Claude Sonnet 4.6 $-$ Qwen\,2.5-32B). Green: Sonnet leads. Red: Qwen leads.}
\label{fig:delta}
\end{figure}

The 0.12 overall gap is not uniform (Figure~\ref{fig:delta}).
Sonnet's advantages concentrate on \textit{cross\_indicator} ($\Delta$\,=\,+0.40), \textit{consistency\_paraphrase} (+0.22), and \textit{regime\_transition} (+0.19): families requiring integration of quantitative signals or robustness to surface variation.
Near-parity on \textit{perturbation} (0.00), \textit{multi\_hop} (+0.01), \textit{fol\_inference} ($-$0.01).
Qwen\,2.5-32B leads decisively on \textit{indicator\_diagnostic} ($-$0.46).

\subsection{Prediction Bias}

\begin{figure}[t]
\centering
\includegraphics[width=0.85\linewidth]{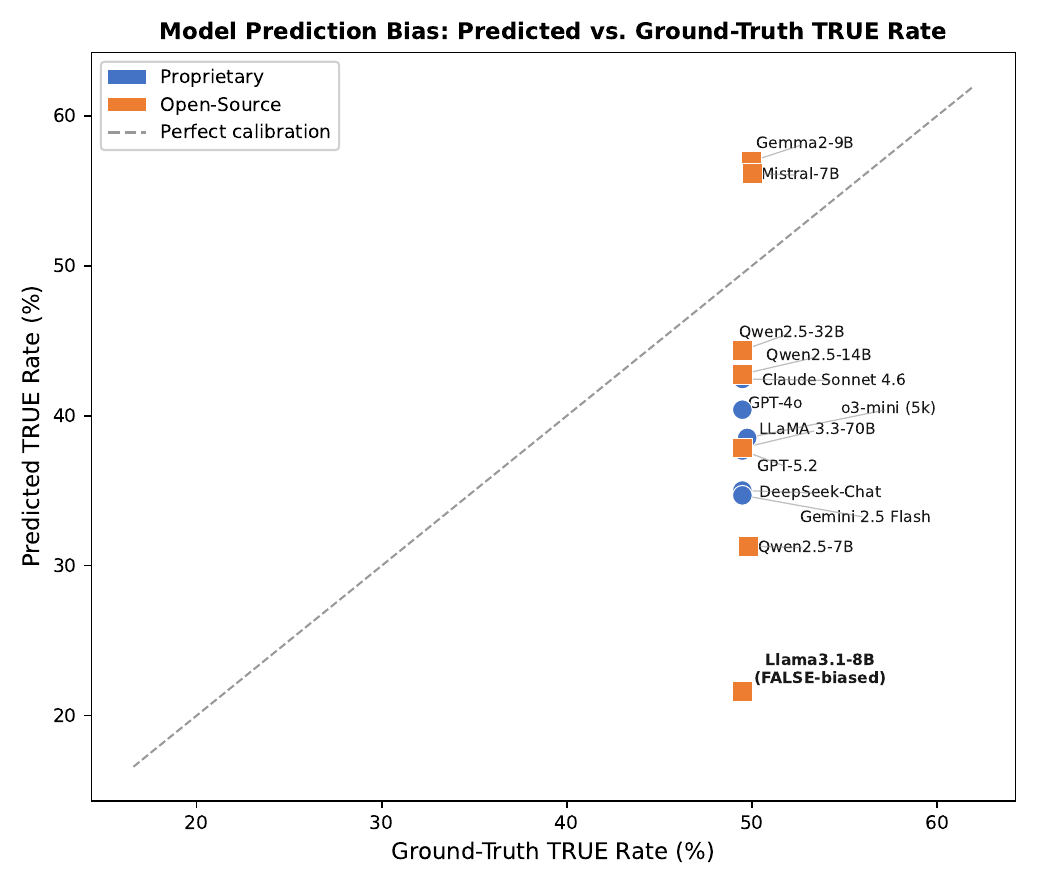}
\caption{Predicted vs.\ ground-truth TRUE rate (49.5\%). LLaMA\,3.1-8B predicts TRUE only 21.6\%.}
\label{fig:bias}
\end{figure}

The ground-truth TRUE rate is 49.5\%.
Most models predict TRUE 42--50\%, but LLaMA\,3.1-8B predicts TRUE only 21.6\% (TNR\,=\,0.88, TPR\,=\,0.32), explaining its low MCC despite moderate balanced accuracy.
Mistral-7B shows the opposite: 56.2\% predicted TRUE (Figure~\ref{fig:bias}).

\section{Discussion}

\paragraph{A knowledge-type boundary.}
The central finding is a dissociation between two types of reasoning.
FOL inference (MCC\,=\,0.52) tests whether models can apply deductive rules when premises are explicitly stated; regime transition (MCC\,=\,0.05) tests whether they can supply numerical premises themselves (e.g., the logistic map transitions to chaos at $r \approx 3.57$).
This is not a scaling problem: within the Qwen\,2.5 family, increasing from 7B to 32B parameters improves multi-hop and FOL inference but leaves regime transition near-random (Appendix~\ref{app:scaling}).
The gap identifies a precise boundary between what LLMs can learn from text (logical rule-following) and what requires numerical grounding (parameter-dependent dynamics).

This complements findings from PrOntoQA~\citep{saparov2023prontoqa}, which showed that LLMs struggle with longer synthetic reasoning chains.
Our results show that even short chains succeed when premises are given (FOL inference), but the difficulty shifts from chain length to premise availability: models cannot generate the quantitative facts needed for bifurcation reasoning.

\paragraph{Consistency failures expose fragile retrieval.}
Consistency\_paraphrase (MCC\,=\,0.25) and perturbation (MCC\,=\,0.26) are not knowledge gaps.
Models answer ``Is Lorenz-63 chaotic?'' correctly at the atomic level (MCC\,=\,0.32--0.62) but flip when the same fact is rephrased.
The knowledge exists; the retrieval is sensitive to surface form.

\paragraph{Proprietary vs.\ open-source: not a monolithic gap.}
The per-family decomposition (Figure~\ref{fig:delta}) shows the gap concentrates on cross-indicator reasoning (+0.40) and consistency (+0.22), while formal deduction and perturbation robustness show near-parity.
Qwen\,2.5-32B's MCC\,=\,0.91 on indicator\_diagnostic (interpreting Lyapunov exponents, permutation entropy, MEGNO) exceeds every proprietary model, suggesting that training data composition matters more than the proprietary/open-source divide for quantitative threshold reasoning.

\paragraph{MaxSAT axiom repair.}
We apply a MaxSAT post-processor (RC2~\citep{ignatiev2019rc2}) that repairs per-system predicate assignments to satisfy all 78 axiom edges with minimal prediction flips.
On atomic questions, repair eliminates all FOL violations and improves MCC substantially for weak models: LLaMA\,3.1-8B gains +0.11 MCC (0.22$\rightarrow$0.33) with 11.2\% of predictions flipped; Mistral-7B gains +0.11 (0.17$\rightarrow$0.28) with 19.0\% flipped.
Claude Sonnet 4.6 is barely affected ($-$0.006 MCC).
When we propagate the repaired predicates to compositional families (multi\_hop, fol\_inference), the picture reverses: repair \emph{degrades} MCC on these families (e.g., $-$0.20 on multi\_hop for LLaMA), because compositional questions encode reasoning that goes beyond per-predicate consistency.
This reveals a separation between two error types: \emph{axiom-inconsistent} errors (fixable by constraint enforcement) and \emph{reasoning} errors (requiring deeper inference).
Solver augmentation helps the first type but not the second, identifying a precise boundary for hybrid LLM-solver approaches.

\paragraph{Chain-of-thought preliminary.}
We tested CoT prompting on the two hardest families (regime\_transition, cross\_indicator; 135 questions) using LLaMA\,3.1-8B.
CoT dramatically increased the invalid rate: 66/68 regime\_transition responses and 57/67 cross\_indicator responses were unparseable (the model generates reasoning text but fails to produce a final TRUE/FALSE).
Among the few parseable CoT responses, regime\_transition remained at MCC\,=\,0.0.
This suggests that for small models, CoT on hard scientific reasoning families introduces a format-compliance problem without solving the underlying reasoning gap.

\subsection{Limitations}

Three families have $N < 100$ (regime\_transition, cross\_indicator, extended\_systems); results on these carry high variance.
The CoT experiment uses only one small model (8B); larger models with CoT may behave differently.
The TRUE/FALSE format cannot distinguish correct reasoning from correct guessing.
Well-known systems (Lorenz-63) may be memorized from pretraining rather than reasoned about; the extended\_systems ceiling effect is consistent with this.

\section{Conclusion}

ChaosBench-Logic v2 and the CARE evaluation protocol together reveal that apparent LLM reasoning performance hides three pathologies: prior collapse (LLaMA\,3.1-8B achieves 60\% accuracy with only 32\% TPR), surface-form fragility (consistency MCC\,=\,0.25), and inability to reason about parameter-dependent dynamics (regime transition MCC\,=\,0.05).
The knowledge-type boundary between rule-following and parametric reasoning does not close with scale.
Our MaxSAT repair experiment reveals two distinct error types: axiom-inconsistent errors (fixable by constraint enforcement, +0.11 MCC on atomic questions for weak models) and reasoning errors on compositional families (degraded by repair), identifying a precise boundary for solver-augmented approaches.
Three directions follow.
First, chain-of-thought evaluation: preliminary results from v1 of this benchmark~\citep{thomas2026chaosbench} found that CoT \emph{decreased} overall accuracy by 2--6 percentage points for both GPT-4 and LLaMA-3, suggesting that explicit reasoning introduces errors on scientific questions where zero-shot retrieval is more reliable.
Whether this pattern holds on v2's harder families (regime\_transition, cross\_indicator) remains an open question.
Second, deeper solver integration: our MaxSAT repair fixes axiom-inconsistent errors but not reasoning errors; coupling LLMs with numerical integrators could address the compositional gap.
Third, fine-tuning on scientific corpora could test whether consistency failures reflect missing knowledge or architectural limitations.
The benchmark, CARE protocol, and released artifacts are publicly available at
\url{https://github.com/11NOel11/ChaosBench-Logic}
and
\url{https://huggingface.co/datasets/11NOel11/ChaosBench-Logic}.

\section*{Reproducibility Statement}

All evaluations use deterministic inference (temperature\,=\,0 where supported; reasoning models are deterministic by design).
Metrics are computed from raw confusion matrices in prediction logs; every number was verified against these logs.
The dataset is generated deterministically from the FOL axiom system.
Code, canonical dataset files, and released artifacts are available at
\url{https://github.com/11NOel11/ChaosBench-Logic}
and
\url{https://huggingface.co/datasets/11NOel11/ChaosBench-Logic}.

\bibliography{references}
\bibliographystyle{iclr2026_conference}

\newpage
\appendix

\section{Full Leaderboard and Subset Evaluations}
\label{app:subset}

\begin{figure}[h]
\centering
\includegraphics[width=\linewidth]{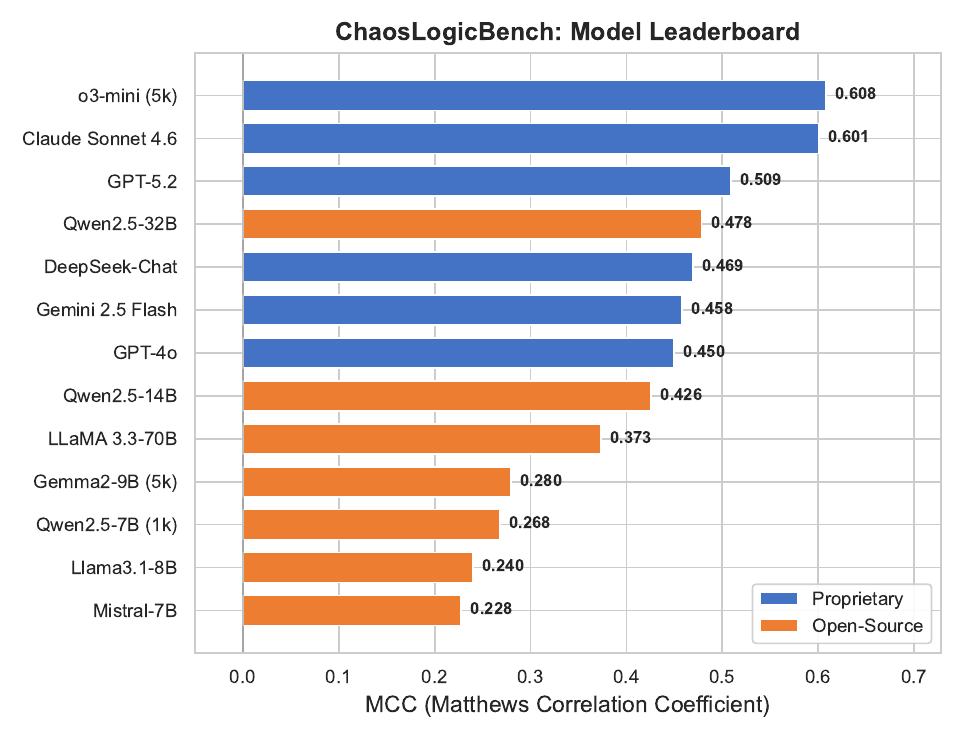}
\caption{MCC by model. Blue: proprietary. Orange: open-source.}
\label{fig:leaderboard}
\end{figure}

\begin{table}[h]
\caption{Subset evaluations. 5k tracks full-canonical MCC within 0.01 (validated on 3 models with both); 1k has higher variance.}
\label{tab:subset}
\centering
\small
\begin{tabular}{llccc}
\toprule
\textbf{Model} & \textbf{Type} & \textbf{Bal.\ Acc} & \textbf{MCC} & \textbf{N} \\
\midrule
o3-mini & Prop. & 0.796 & 0.608 & 5,000 \\
Gemma2-9B & OSS & 0.639 & 0.280 & 5,000 \\
GPT-4o-mini & Prop. & 0.623 & 0.272 & 1,000 \\
Qwen\,2.5-7B & OSS & 0.624 & 0.268 & 1,000 \\
\bottomrule
\end{tabular}
\end{table}

\section{Model Size Scaling}
\label{app:scaling}

\begin{figure}[h]
\centering
\includegraphics[width=0.85\linewidth]{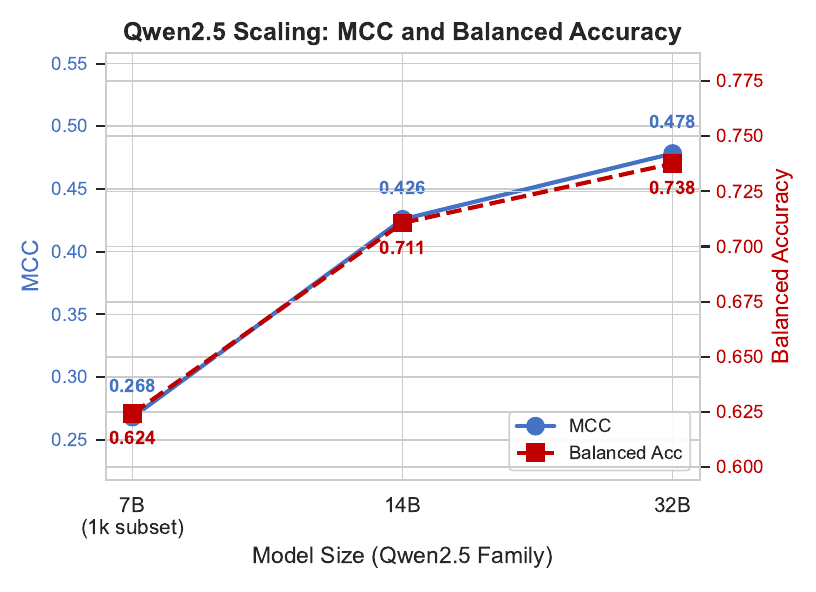}
\caption{Scaling within Qwen\,2.5. MCC increases from 0.27 (7B, 1k subset) to 0.48 (32B), but regime\_transition remains near-random at all scales.}
\label{fig:scaling}
\end{figure}

\section{Subset Validation}
\label{app:crosscheck}

\begin{figure}[h]
\centering
\includegraphics[width=0.85\linewidth]{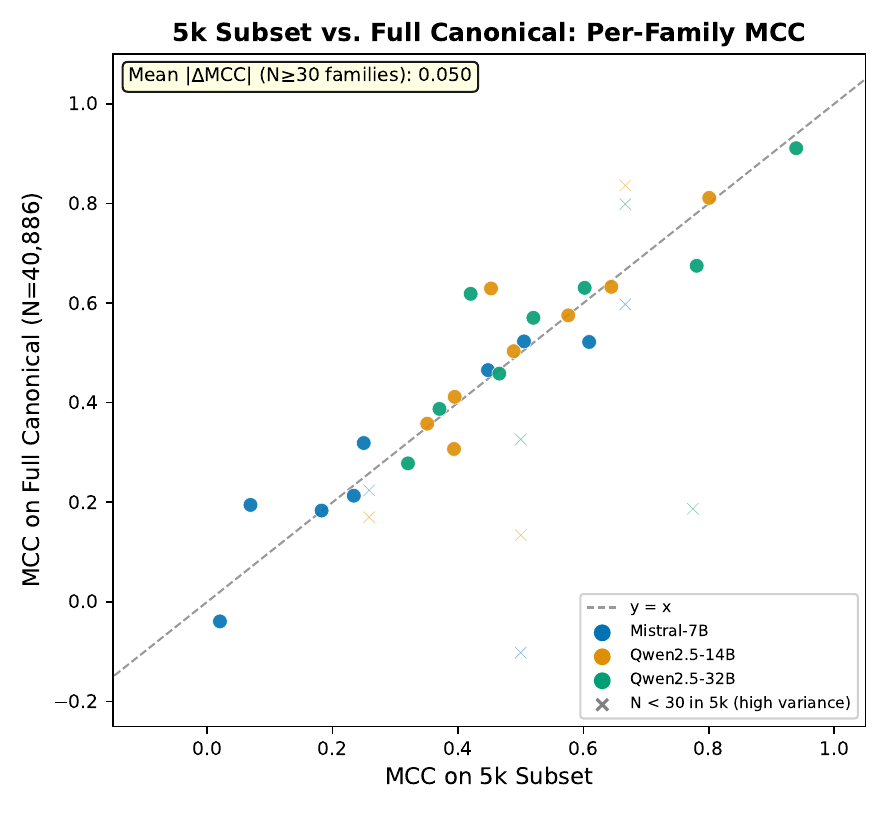}
\caption{5k vs.\ full canonical per-family MCC. Circles: $N \geq 30$. Crosses: $N < 30$ (high variance).}
\label{fig:crosscheck}
\end{figure}

For models with both evaluations (Qwen\,2.5-32B, Qwen\,2.5-14B, Mistral-7B), mean $|\Delta\text{MCC}| < 0.01$ overall; per-family mean $|\Delta\text{MCC}| = 0.050$ for families with $N \geq 30$ in the 5k subset.

\section{Family Discrimination}
\label{app:discrimination}

\begin{figure}[h]
\centering
\includegraphics[width=0.85\linewidth]{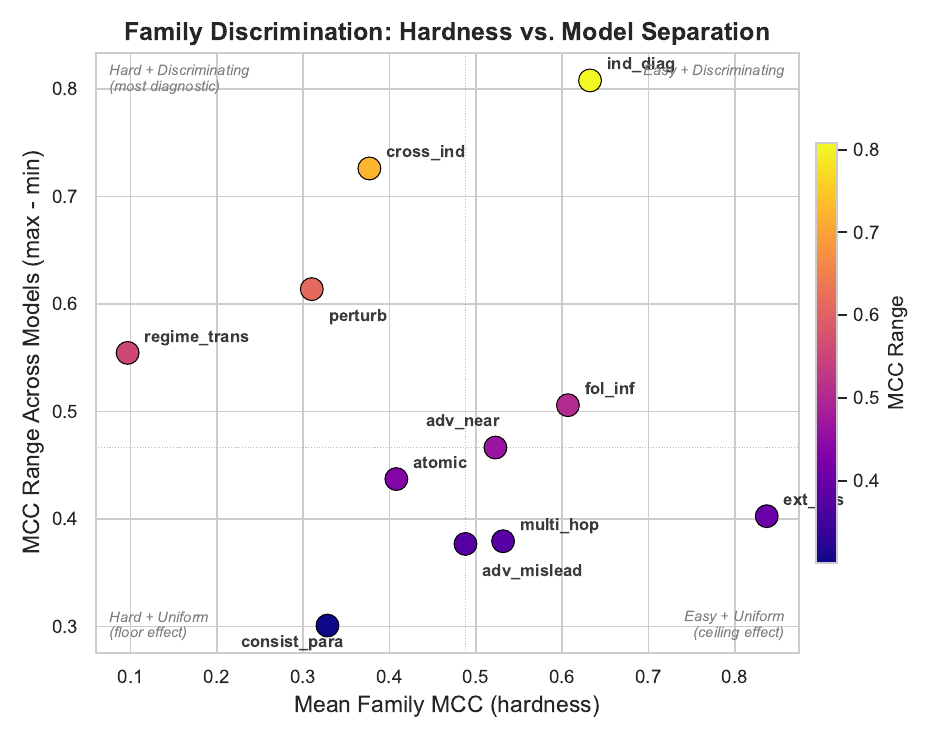}
\caption{Mean MCC vs.\ MCC range across 10 models. Upper-left: hard and discriminating. Lower-left: floor effects.}
\label{fig:discrimination}
\end{figure}

\section{Question Examples}
\label{app:examples}

\begin{table}[h]
\caption{Example questions with ground truth (GT) and predictions. Son.\,=\,Claude Sonnet 4.6; Qw.\,=\,Qwen\,2.5-32B; Ll.\,=\,LLaMA\,3.3-70B. Incorrect predictions in \textbf{bold}.}
\label{tab:examples}
\centering
\small
\begin{tabular}{p{1.2cm}p{7.0cm}cccc}
\toprule
\textbf{Family} & \textbf{Question (abbreviated)} & \textbf{GT} & \textbf{Son.} & \textbf{Qw.} & \textbf{Ll.} \\
\midrule
multi\_hop & ``FluidTrampoline is strongly mixing. Strongly mixing $\Rightarrow$ weakly mixing $\Rightarrow$ ergodic $\Rightarrow$ bounded. Is it bounded?'' & T & T & \textbf{F} & \textbf{F} \\
\addlinespace
regime & ``At $\alpha$=15.6, is Chua's circuit chaotic?'' & T & T & T & T \\
\addlinespace
fol\_inf & ``Given Torus is stat.\ predictable, must it be bounded?'' & T & T & T & \textbf{F} \\
\addlinespace
adv\_misl & ``Given bounded, can R\"ossler be weak mixing?'' & T & T & \textbf{F} & \textbf{F} \\
\bottomrule
\end{tabular}
\end{table}

\section{Axiom Specification}
\label{app:axioms}

The six primary regime axioms:
\begin{enumerate}[nosep]
\item \textbf{Chaotic} $\Rightarrow$ Deterministic $\land$ PosLyap $\land$ Sensitive $\land$ PointUnpredictable $\land$ StatPredictable $\land$ Mixing; excludes Random, Periodic, QuasiPeriodic, FixedPointAttr.
\item \textbf{Random} excludes Deterministic, Chaotic, QuasiPeriodic, Periodic.
\item \textbf{QuasiPeriodic} $\Rightarrow$ Deterministic $\land$ Bounded; excludes Chaotic, Random, Periodic, FixedPointAttr.
\item \textbf{Periodic} $\Rightarrow$ Deterministic $\land$ Bounded; excludes Chaotic, Random, QuasiPeriodic, StrangeAttr.
\item \textbf{FixedPointAttr} $\Rightarrow$ Deterministic; excludes Chaotic, Random, QuasiPeriodic, Periodic, StrangeAttr.
\item \textbf{Deterministic} excludes Random.
\end{enumerate}

Additional edges: PosLyap $\Rightarrow$ Sensitive $\Rightarrow$ PointUnpredictable; StrangeAttr $\Rightarrow$ Dissipative $\land$ Bounded; Mixing $\Rightarrow$ Ergodic $\Rightarrow$ Bounded; HyperChaotic $\Rightarrow$ Chaotic $\land$ StrangeAttr $\land$ Dissipative; Conservative $\Rightarrow$ Bounded $\land$ Ergodic; StrongMixing $\Rightarrow$ WeakMixing $\Rightarrow$ Ergodic; ContinuousTime $\leftrightarrow \neg$DiscreteTime; Forced $\leftrightarrow \neg$Autonomous; DelaySystem $\Rightarrow$ ContinuousTime.

\section{Confusion Matrices}
\label{app:confusion}

\begin{table}[h]
\caption{Regime\_transition (N\,=\,68) confusion matrices.}
\label{tab:confusion}
\centering
\small
\begin{tabular}{lcccc|cc}
\toprule
\textbf{Model} & \textbf{TP} & \textbf{FP} & \textbf{TN} & \textbf{FN} & \textbf{MCC} & \textbf{Bal.\ Acc} \\
\midrule
LLaMA\,3.3-70B & 9 & 17 & 20 & 22 & $-$0.173 & 0.415 \\
Mistral-7B & \multicolumn{4}{c|}{\textit{(from aggregate)}} & $-$0.102 & -- \\
Claude Sonnet 4.6 & 15 & 5 & 32 & 16 & +0.381 & 0.674 \\
\bottomrule
\end{tabular}
\end{table}

\section{Invalid Rates}
\label{app:invalid}

Most models produce zero invalids. Mistral-7B has the highest rate at 1.1\%; LLaMA\,3.1-8B $<$0.01\%; all others 0.0\%.

\section{Prompt Template}
\label{app:prompt}

\begin{quote}
\small
\texttt{Answer the following question about the dynamical system. Reply with only TRUE or FALSE.}\\
\texttt{Question: [question text]}\\
\texttt{Answer:}
\end{quote}

No system prompt, few-shot examples, or CoT instructions.
Temperature\,=\,0 for all models that accept the parameter. Reasoning models (o3-mini, GPT-5.2) do not accept a temperature argument; their outputs are internally deterministic via the reasoning process.
Max tokens: 16 for most models; 1024 for o3-mini and GPT-5.2 (reasoning token budget), and Gemini 2.5 Flash (thinking process).

\end{document}